\documentclass[letterpaper, 10 pt, conference]{ieeeconf}  % Comment this line out

\makeatletter
\let\NAT@parse\undefined
\makeatother

\usepackage{amsmath} % assumes amsmath package installed
\usepackage{amssymb}  % assumes amsmath package installed
\usepackage{bm}
\usepackage[pdftex]{graphicx}
\usepackage[numbers, square, comma, sort&compress]{natbib}

\IEEEoverridecommandlockouts                              % This command is only
\title{\LARGE \bf
Resolving Implementation Ambiguity and Improving SURF
}

\author{Peter Abeles% <-this % stops a space
\thanks{Prepublication draft v3 \;\;{\tt\small peter.abeles@gmail.com}}% <-this % stops a space
}

\begin{document}

\maketitle
\thispagestyle{empty}
\pagestyle{empty}

%%%%%%%%%%%%%%%%%%%%%%%%%%%%%%%%%%%%%%%%%%%%%%%%%%%%%%%%%%%%%%%%%%%%%%%%%%%%%%%%
\begin{abstract}
Speeded Up Robust Features (SURF) has emerged as one of the more popular feature descriptors and detectors in recent years.  Performance and algorithmic details vary widely between implementations due to SURF's complexity and ambiguities found in its description. To resolve these ambiguities, a set of general techniques for feature stability is defined based on the smoothness rule. Additional improvements to SURF are proposed for speed and stability. To illustrate the importance of these implementation details, a performance study of popular SURF implementations is done. By utilizing all the suggested improvements, it is possible to create a SURF implementation that is several times faster and more stable.
\end{abstract}

%%%%%%%%%%%%%%%%%%%%%%%%%%%%%%%%%%%%%%%%%%%%%%%%%%%%%%%%%%%%%%%%%%%%%%%%%%%%%%%%
\section{Introduction}

Image correspondence is the problem of associating features inside one image against another related image. Knowing image correspondence allows for the scene's structure and camera motion to be determined, in addition to object recognition. For point based image correspondence, a typical processing flow includes detecting interest points, describe regions, and feature association.

In recent years, Speeded-Up Robust Features (SURF) \cite{Bay2008} has emerged as a popular choice for interest point detection and region description. Building upon previous work (e.g. SIFT \cite{Lowe2004}), SURF is primarily designed for speed and invariance in scale and in-plane rotation. While skew, anti-isotropic scaling, and perspective effects are considered second order.

SURF's performance is often used as a benchmark, against which other recently developed feature detectors are measured \cite{Tola2009,Luo2009,Calonder10}. However, which implementation of SURF should be used for these comparisons? SURF is not trivial to implement and there are many options from which to choose.  A binary reference implementation \cite{Reference} has been provided without source code by the original author, but as of publication, is not compatible with the latest distributions of Linux. As has been noted in \cite{Agrawl2008,Gossow2011}, ambiguities exist in the original paper and the reference binary runs slower than expected.

To create a stable region descriptor, small changes in location and scale must cause a proportionally small change in descriptor value. This observation will be referred to as the \emph{smoothness rule}.  Similar statements are made by D. Lowe \cite{Lowe2004} and justified by a biological vision model \cite{Edelman1997}.  How to enforce the smoothness rule is not well understood in every situation, and is often incorrectly applied.

The primary point of contention regarding SURF is the interpolate method when computing the descriptor. This ambiguity has led to several different interpretations as well as proposals for improving SURF \cite{Agrawl2008,OpenSURF,BoofCV,PanOMatic}. Other important details, such as how to handle image borders, are also never discussed. To resolve these ambiguities, general techniques are proposed for enforcing the smoothness rule and then applied to different components of SURF. Additional new techniques are proposed for improving SURF's speed and stability.

This work seeks to address ambiguities found in the original paper, explore simple ways to improve performance, and compare popular implementations of SURF.  Different SURF implementations are compared based upon descriptor stability, detector stability, and runtime speed.  Many performance studies comparing different descriptors have been done in the past.  One recent study \cite{Gossow2011} focused only on open source SURF implementations, but had a smaller scope in terms of implementations and discussion than this work.  The purpose of this performance study is to highlight the importance of low level implementation details and to identify which implementations are best used to characterize SURF's performance.

%%%%%%%%%%%%%%%%%%%%%%%%%%%%%%%%%%%%%%%%%%%%%%%%%%%%%%%%%%%%%%%%%%%%%%%%%%%%%%%%
\section{Speeded-Up Robust Features}
\label{section:SURF}

The following is a high level overview of the SURF detector and descriptor.  For a complete discussion consult the SURF paper \cite{Bay2008}.   SURF achieves speed across a range of scales through the use of integral images \cite{Viola2001,Simard1998}.  Transforming an image into an integral image allows the sum of all pixels contained inside arbitrary axis aligned rectangle to be found in four operations.

The value of each pixel $(x,y)$ in the integral image $I_{\Sigma}$ is computed by summing pixel intensities within a rectangle up to $(x,y)$:
\begin{equation}
I_{\Sigma}(x,y) = \sum_{i=0}^{i\le x}\sum_{j=0}^{j\le y} I(i,j)
\end{equation}
Then to find the sum of pixel values contained in a rectangle $R_{\Sigma}$ compute:
\begin{equation}
\begin{array}{c}
R_{\Sigma}(x_1,y_1,x_2,y_2) = I_{\Sigma}(x_2,y_2) - I_{\Sigma}(x_2,y_1-1) \\
- I_{\Sigma}(x_1-1,y_2) + I_{\Sigma}(x_1-1,y_1-1)
\end{array}
\end{equation}
where $(x_1,y_1) \le (x_2,y_2)$.

Interest point detection is done using an approximation of the Hessian determinant scale-space detector \cite{Lindeberg1998}.  The Hessian's determinant is found by approximating the Gaussian's second order partial derivatives ($D_{xx},D_{yy},D_{xy}$) using box integrals, as described in \cite{Bay2008}.  
\begin{equation}
det(H) = D_{xx}D_{yy} - (wD_{xy})^2
\end{equation}
This is done across different sized regions and scales.  Interest points are defined as local maximums in the 2D image and across scale space.  Scale and location are interpolated by fitting a 3D quadratic \cite{Brown2002} to feature intensity values in the local 3x3x3 region.

Several different variations on the SURF descriptor are described in \cite{Bay2008}, but only the oriented SURF-64 descriptor is considered in this study.  Orientation is estimated by computing the gradient\footnote{In Bayes \emph{et al.} \cite{Bay2008} the gradient operator above is referred as Haar wavelets or $(d_x,d_y)$.  While it is true they are Haar wavelet like, it is the opinion of this author that invoking wavelet theory causes more confusion than insight.  Some might consider it more intuitive to think of these as gradient operators adjusted for scale.} inside a neighborhood of radius of $6s$, where $s$ is the feature's scale.  The gradient is weighted by a Gaussian centered at the interest point, its angle computed, and saved into an array.  Using a moving window of $\frac{\pi}{3}$ radians, the window with the largest gradient sum is found and the feature's orientation computed from its sum.

The feature description is computed inside a square region of size $20s$, aligned to the found orientation.  This region is then broken up into a 4 by 4 grid for a total of 16 subregions, which are of size $5s$.  For each subregion the sum of the gradient and sum of the gradient's absolute value is computed:
\begin{equation}
v = (\sum d_x , \sum d_y , \sum |d_x| , \sum |d_y|)
\end{equation}
These responses are weighted using a Gaussian distribution.  Each subregion contributes 4 features ($v$), resulting in a total of 64 features for the descriptor.

The gradient can only be efficiently computed along the image's axis.  To accommodate for the feature's orientation, the gradient is rotated so that it is oriented along the feature's axis.

%%%%%%%%%%%%%%%%%%%%%%%%%%%%%%%%%%%%%%%%%%%%%%%%%%%%%%%%%%%%%%%%%%%%%%%%%%%%%%%%
\section{Implementation Details}

To create a stable region descriptor, the smoothness rule discussed in the introduction must be enforced.  The following are several general techniques for enforcing the smoothness rule:  1) Use continuous interpolation functions when sampling image intensity values.  2) Increase a sample region's size to improve stability by reducing the fractional change in value when crossing a pixel border.  3) Avoid interacting with image and object borders. 4) Maintain a constant value when interacting with image borders.  Technique 2 and 3 can be conflicting since as the region size increases it is more likely to interact with the boundary conditions.  

By applying these general techniques to SURF, important implementation details that were omitted or ambiguously described are resolved.  In addition, new approaches for improve the speed and stability of SURF are presented in this section.

\subsection{Descriptor Interpolation}

Interpolation of the gradient's response when computing $v$ is not fully described in \cite{Bay2008}.  This has resulted in several different algorithmic interpretations.  The most straight forward interpretation is to use nearest neighbor interpolation. However, this method does not have a smooth transition between pixel boundaries, degrading descriptor stability.

Agrawl \emph{et al} \cite{Agrawl2008} propose to have each subregion overlap by adding a padding of 2s and to weigh the gradient using a subregion centered Gaussian distribution.  The resulting descriptor has a region of size $24s$.  Pan-o-Matic \cite{PanOMatic} samples the gradient using a variable number of points depending on the ratio of region size to sample size and then uses bilinear interpolation to compute the descriptor values.  The Pan-o-Matic interpolation technique is similar to how interpolation is done in SIFT.

Overlapping subregions and bilinear interpolation produced similar stability performance when given the same inputs.  However, overlapping subregions lend themselves towards a faster and easier implementation.  Nearest neighbor interpolation is the fastest and is stable enough for many applications.

\subsection{Image Border}

Interest points near borders can have descriptors whose region goes outside of the image.  Detected interest point corresponds to a region of size $1.2*9s$ while a descriptor covers a region of $20s.$  The SURF paper does not specify how to handle pixels outside of the image.

Several different techniques for handling image borders have been observed. A) Treat all pixels outside of the image as having a value of zero. B) Setting the response of any operator crossing the image border to be zero. C) Extending the image using the closest edge pixel value. D) Extending the image using reflection. E) Discarding features which intersect the image border.

The best approach seen in practice is B, but C could produce better results.  When using approach B there is an abrupt change in value once an operator cross the border, but once any part is outside its value stays constant.  Approach C would not have an abrupt value change at the border and would converge towards a constant value as the operator moves outside the image.

The same cannot be said for the other approaches. A) Operators converge towards zero using a stepwise function as they move further out of the image.  D) Values of operators do not converge and constantly change. E) Throws away too many useful features that can be reliably associated.

The approach which best follows the smoothness rule is C, but none of the implementations considered used that approach.  Approach D is not used by any of the evaluated implementations, but is used by \cite{IPOL:SURF2011}.

\subsection{Interest Point Interpolation}

After an interest point has been detected using non-maximum suppression, its position $(x,y,s)$ is interpolated as the extreme of a 3D quadratic.  The procedure described by Brown and Lowe \cite{Brown2002} uses the Laplacian computed with pixel differences.  Estimating second order derivatives using pixel differences amplifies noise and is further approximated using box integrals.  Ad hoc modifications are required to filter out illogical solutions generated with this approach.

To avoid these issues, a quadratic can instead be fit directly to sampled intensity values.  If the minimum number of points are used and the center point is the peak, the interpolated peak must lie inside the sample region.  An approach used in BoofCV \cite{BoofCV} fits quadratic 1D polynomials across each $(x,y,s)$ axis independently.  While not capturing off axis structural information, it is more stable and requires fewer operations.

\subsection{Coordinate Discretation}

Sampling coordinates do not align with integer image coordinates during descriptor computations because the region is scaled and rotated.  To minimize the expected error, the round operator should be used when discretizing.  Casting to an integer is equivalent to flooring (all image coordinates are positive), which has a larger expected error than round.  To improve runtime performance the round operator should not be used directly.  Instead coordinates should add 0.5 then be cast into an integer.  Often, adding 0.5 only needs to be done once per axis inside an image processing loop.

\subsection{Derivative Operator}

\begin{figure}
\center
\includegraphics[trim = 0mm 125mm 120mm 0mm, clip, scale=0.6]{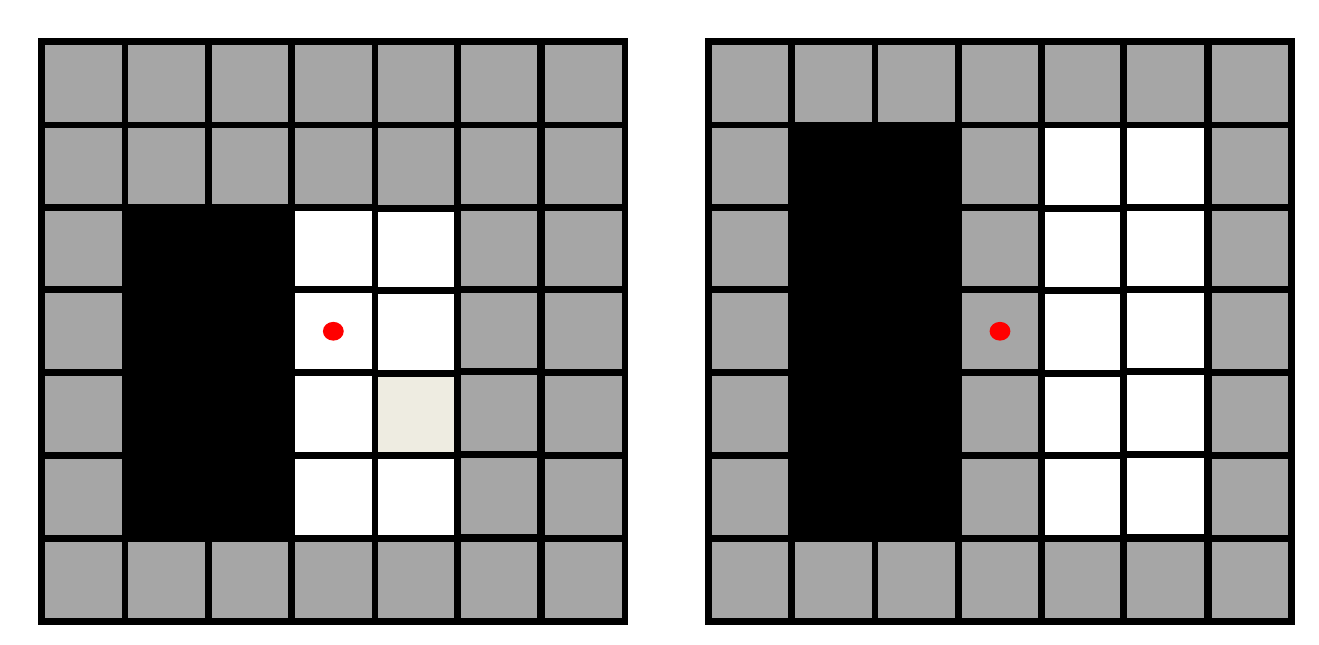}
\caption{\label{fig:gradient}Left: original Haar-like derivative operator. Right: proposed symmetric derivative operator. Red dot indicates the region's center. The Haar kernel lacks an obvious center and is not symmetric, causing a bias.  Dark squares indicate a weight of -1 and white squares +1.}
\end{figure}

The Haar wavelet like derivative kernel used in SURF lacks symmetry about the sampled pixel.  The lack of symmetry creates a bias and it is ambiguous which pixels is the center.  An alternative symmetric derivative operator is proposed that overcomes these issues, see Figure \ref{fig:gradient}.  The alternative kernel has a width of $w=\mbox{rnd}(2rs)+1$, where $r$ is the radius at a scale of one.  A value of $r=1$ is recommended for descriptor computations.

\subsection{Laplacian Sign}

Another smaller performance boost can be found in delaying the Laplacian's sign computation.  It is stated in \cite{Bay2008} that the Laplacian's sign can be computed with no loss in performance.  This is not quite true; the computation requires an additional operation for each pixel and scale, plus storage.  Instead if the Laplacian sign is computed for found interest points only, then 24 additional operations are required per feature.  Since the number of pixels is much greater than the number of found features, the latter is many times faster and requires no additional storage.

\subsection{Orientation Estimation}

The proposed region orientation algorithm in \cite{Bay2008} is computationally expensive, see Section \ref{section:SURF} for a summary.  An alternative and much faster approach is to compute a weighted sum of the gradient $(\sum d_y,\sum d_x)$ and then find angle using $\mbox{atan2}(\sum d_y,\sum d_x)$.  However, the improved speed comes at the cost of some stability.

\subsection{Inner Loop Optimization}

One common technique (often ignored) for improving performance is to optimize the inner image processing loops.  The easiest and most straight way to write image processing code is to write a single function that iterates through each image pixel and checks for boundary conditions.  The disadvantage of this approach is that it forces a check that is unnecessary on the vast majority of image pixels and makes it more difficult for a compiler to optimize the code.  Instead two functions should be written, one which only processes the border and a second which is highly optimized for processing the inner image.

\subsection{Tuning Parameters}

All implementations deviated from recommended turning parameter values found in \cite{Bay2008}.  More successful implementations used larger kernels or regions when sampling the image.

%%%%%%%%%%%%%%%%%%%%%%%%%%%%%%%%%%%%%%%%%%%%%%%%%%%%%%%%%%%%%%%%%%%%%%%%%%%%%%%%
\section{Test Setup}
\begin{figure*}
\centering
\begin{tabular}{|l|c|c|c|c|}
\hline
Implementation & Cite & Version & Language & Comment \\
\hline
BoofCV-F &\cite{BoofCV} & v0.5 & Java & Faster but less accurate \\
\hline
BoofCV-M &\cite{BoofCV} & v0.5 & Java & Slower but more accurate \\
\hline
JavaSURF &\cite{JavaSURF} & SVN r4 & Java & No orientation \\
\hline
JOpenSURF &\cite{JOpenSURF} & SVN r24 & Java & Java port of OpenSURF \\
\hline
OpenCV &\cite{OpenCV} & 2.3.1 SVN r6879 & C++ &  \\
\hline
OpenSURF &\cite{OpenSURF} & 27/05/2010 & C++ & \\
\hline
Pan-o-Matic &\cite{PanOMatic} & 0.9.4 & C++ &  \\
\hline
Reference &\cite{Reference} & 1.0.9 & C++ & Provided by original author \\
\hline
\end{tabular}
\caption{\label{librarylist}List of evaluated implementations in alphabetical order. If a formal version is lacking or insufficient then a repository version is referenced.}  
\end{figure*}

Evaluation is performed using test image sequences from Mikolajczyk and Schmid \cite{Mikolajczyk2005}.  Each sequences has a set of known image homographies relating images to the first image in the sequence.  Each sequence is designed to test different types of distortion and image noise.   The evaluated data sets include ``bark'', ``bikes'', ``boat'', ``graf'', ``leuven'', ``trees'', ``ubc'', and ``wall''.

Evaluated library are listed in Table \ref{librarylist}.  Only single threaded libraries are considered.  Both C++ and Java are popular languages for computer vision, with C/C++ being the most popular.  Many other SURF implementations are available and can be easily found online.  There are several multi-threaded and hardware specific (e.g. FPGA, GPU) implementations to choose from.

Additional implementation details: OpenSURF, JOpenSURF, and BoofCV-M all implemented the modified descriptor from \cite{Agrawl2008}.  Pan-o-Matic uses bilinear interpolation when computing the descriptor. BoofCV-F and OpenCV use nearest neighbor interpolation.  Image border technique (A) is used by OpenSURF, JOpenSURF, and JavaSURF, and (B) OpenCV, Pan-o-Matic, BoofCV-F, and BoofCV-M.  The modified derivative is used by both BoofCV implementations. JOpenSURF is a straight forward port of OpenSURF.  JavaSURF lacks the ability to estimate orientation.  

Two variants of BoofCV's descriptor are included in this study.  BoofCV-M uses all recommended techniques that maximize descriptor stability.  BoofCV-F maximizes speed by trading off some stability.    BoofCV only has one detector implementation.

%%%%%%%%%%%%%%%%%%%%%%%%%%%%%%%%%%%%%%%%%%%%%%%%%%%%%%%%%%%%%%%%%%%%%%%%%%%%%%%%
\section{Performance Metrics}

Standard performance metrics are used to evaluate detector stability, descriptor stability, and runtime speed.  Performance for runtime speed is measured as elapsed time.  Performance metrics for the descriptor and detector stability are described in the following sub-sections.

\subsection{Descriptor}
Descriptor stability is measured based on the fraction of correct associations.  Even though the true locations of interest points are known, approximate locations from a detector are used instead.  Exact locations are not realistic and a good descriptor needs to handle small errors in location.

Two features are associated if they are mutually each other's best match using Euclidean error.  An association is declared as being correct if the matching pair is within three pixels of the truth.

Summary statistics shown in Figure \ref{fig:overall_descriptor} are found for each implementation by summing the fraction of correct associations across each image in every sequences and dividing by best implementation's score.

\subsection{Detector}

Detector stability is measured using repeatability \cite{Schmid2000}, which ``signifies that detection is independent of changes in imaging conditions''.   One problem with repeatability is it favors detectors that detect more features \cite{Gauglitz2011}.  Excessive detections increase computational cost without improving association quality.  An extreme example is if every pixel is marked as an interest point, it would have perfect repeatability.

Attempts made to have each implementation detect the same number of interest points across all images proved to be futile.  Some implementations detected an excessive number of points in some but not all images.  To compensate for this issue, the definition of repeatability has been modified to ignore regions with closely packed points.  By only considering interest points with unambiguous matches, repeatability bias is reduced.

The modified repeatability measure $r_i$ is defined below:
\begin{equation}
r_i = \frac{|A_i|-|T_i|}{|P_i|-|T_i|}
\end{equation}
where  $P_i$ is the set all points, $A_i$ is the set of actual matches, and $T_i$ is the set of ignored matches.
\begin{eqnarray}
P_i &=& \{x_a \in F_o | H_i x_a \in I_i \} \\
A_i &=& \{x_c \in F_i | \parallel H_i x_b - x_c \parallel < \epsilon , x_b \in P \} \\
T_i &=& \{x_d \in F_i | \parallel x_c - x_d \parallel < \epsilon , x_d \ne x_c \}
\end{eqnarray}
where $I_i$ is image $i$, $H_i$ is the homography transform from image 1 to $i$, $F_i$ is the set of all detected interest points, $x$ is an interest point, and $\epsilon$ is the match tolerance.  

Two interest points are considered a match if their position and scale are within tolerance.  The true position is found using the provided homography.  Scale is computed by 1) sampling four evenly spaced points one pixel away from the interest point, 2) applying homography transform to each sample point and interest point, 3) finding the distance of transformed sample points from transformed interest point, and 4) setting expected scale to average distance.

Summary statistics shown in Figure \ref{fig:overall_detector} are found for each implementation by summing repeatability across each image in every sequence and dividing by the best implementation's score.

%%%%%%%%%%%%%%%%%%%%%%%%%%%%%%%%%%%%%%%%%%%%%%%%%%%%%%%%%%%%%%%%%%%%%%%%%%%%%%%%

\section{Test Procedure}

Test procedures for descriptor stability, detector stability, and runtime performance are presented below.

\subsection{Descriptor stability}
  
Descriptor stability is measured by computing the description at interest points selected by the reference library.  Each library is configured to describe SURF-64 features.
\begin{enumerate}
\item For each image, detect interest points using the reference detector, save position and scale to a file.  
\item For each library, image, and interest point, compute the region's orientation and create a descriptor.
\item For each library and each sequence, count the number of correct associations between the first image and the $N^{th}$ image
\end{enumerate}

The same detection configuration for all image sequences.  The number of detected features varied by image and ranged from about 1,200 to 10,000.

\subsection{Detector stability}

Interest points are detected for all images in every sequence by each library.  Tuning each library to detect the same number of features in all images proved to be impossible.  Instead they are tuned to detect about 2,000 features in image 1 in the graf sequence.  To compensate for implementations that detected an excessive numbers of features the definition of repeatability is modified, as described above.  

Detector configuration:
\begin{enumerate}
\item 3x3 non-max region
\item Octaves: 4
\item Scales: 4
\item Base Size: 9
\item Pixel Skip: 1
\end{enumerate}

 Tolerance for position is 1.5 pixels and 0.25 for scale.  Relative ranking was found to be insensitive to reasonable changes (e.g. 3 pixels or 0.5 scale) in thresholds.

\subsection{Runtime Speed}

Runtime performance is measured by having each library detect and describe features inside an image.  Detector and descriptor configurations are the same as above.  Evaluation procedure:
\begin{enumerate}
\item Kill all extraneous processes. 
\item Measure elapsed time to detect and describe features.
\item Repeat 10 times in the same process and output best result.
\item Run the whole experiment 11 times for each library and record the median time.
\end{enumerate}

All tests are performed on an desktop computer with Ubuntu 10.10 installed and an Intel Q6600 2.4GHz CPU.  Native libraries are compiled using g++ 4.4.5 with the -O3 flag.  Java libraries are compiled and run using Oracle JDK 1.6.30 64 bit.  No additional flags are passed to the Java Runtime Environment, the -server flag is implicit.  

Native library runtime speeds are highly dependent upon the level of optimization done by the compiler and which instructions they are allowed to use.  For example, Pan-o-Matic runs about three times slower if no optimization flags are specified.  To provide more general performance, additional hardware specific flags are not manually injected into build scripts.

Elapsed time is measured in the actual application using System.currentTimeMillis() in Java and clock() in C++. Java libraries tended to exhibit more variation than native libraries and a short warm up period.

%%%%%%%%%%%%%%%%%%%%%%%%%%%%%%%%%%%%%%%%%%%%%%%%%%%%%%%%%%%%%%%%%%%%%%%%%%%%%%%%

\section{Performance Results}

\begin{figure}[h!]
\centering
\includegraphics[clip=true,trim=2.5cm 9.5cm 1.9cm 9.6cm,scale=0.5]{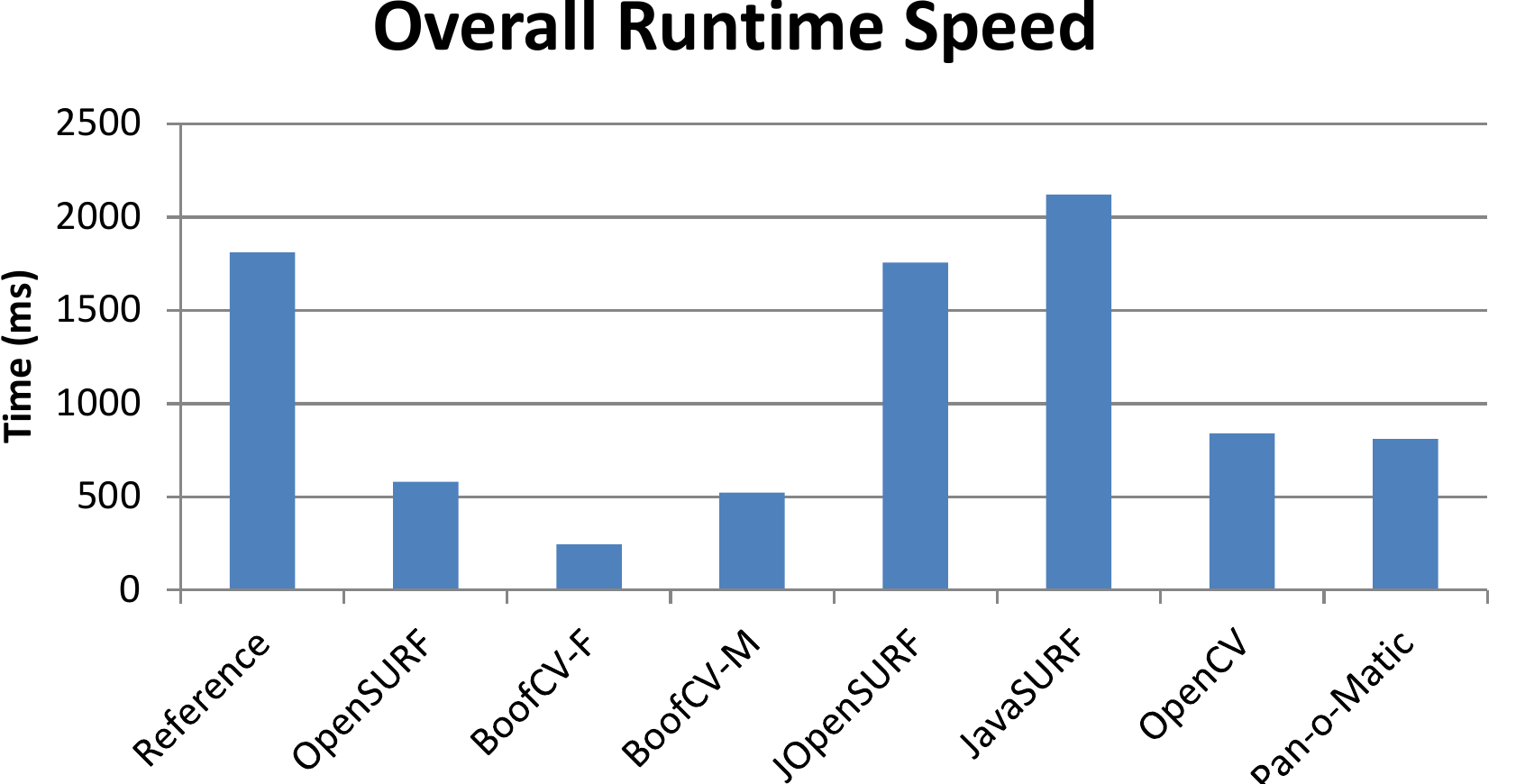}
\caption{\label{fig:overall_runtime}Runtime speed for detecting and describing an image features.  Lower bars are better.  Each library is tuned to detect approximately 2000 features in a 850x680 image.}
\end{figure}

\begin{figure}[h!]
\centering
\includegraphics[clip=true,trim=2.5cm 9.4cm 1.9cm 9.6cm,scale=0.5]{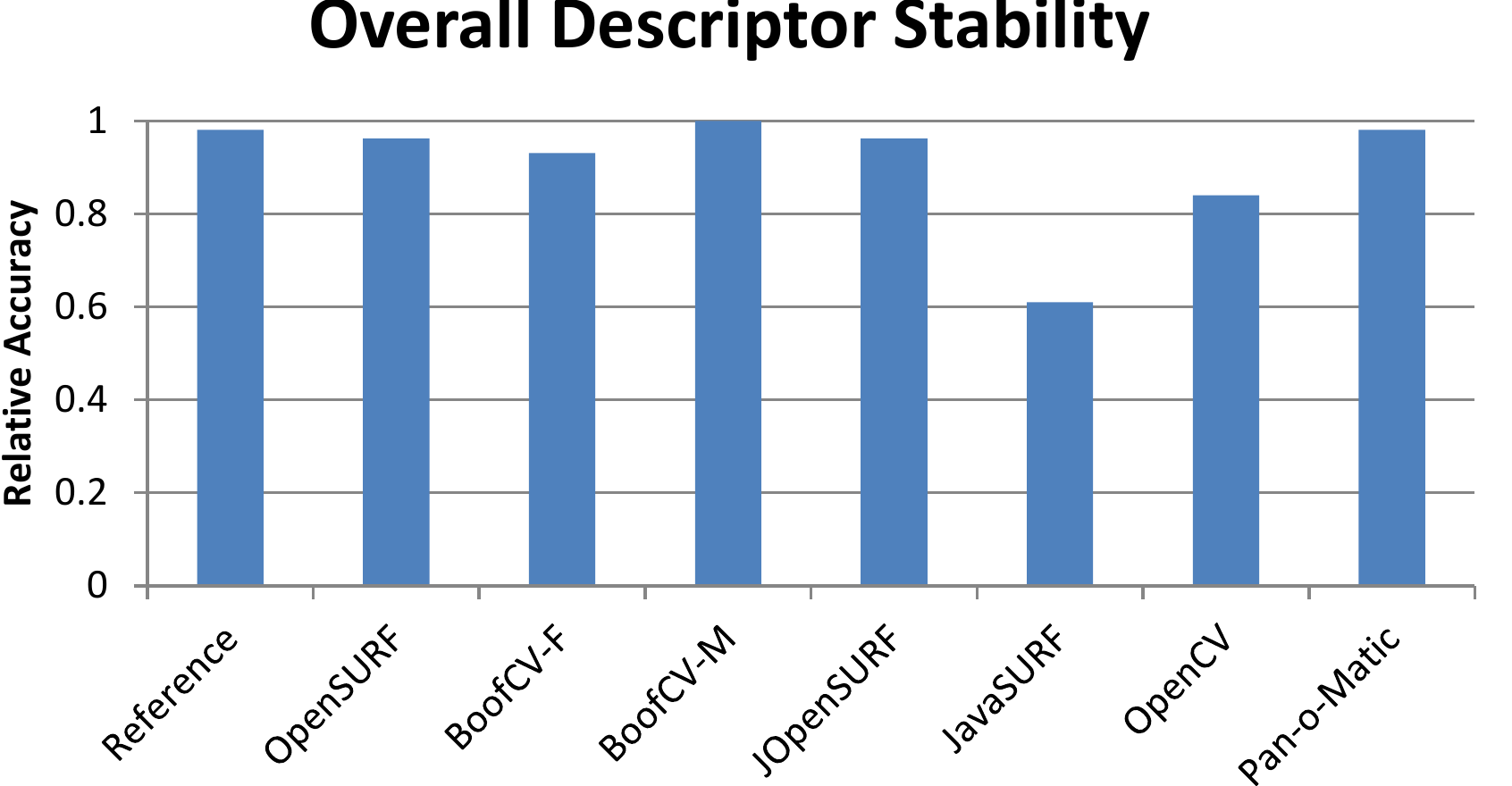}
\caption{\label{fig:overall_descriptor}Summary of descriptor stability using correct association fraction across all image sequences.  Higher bars are better.  Interest points are generated by reference library.}
\end{figure}

\begin{figure}[h!]
\centering
\includegraphics[clip=true,trim=2.5cm 9.5cm 1.9cm 9.6cm,scale=0.5]{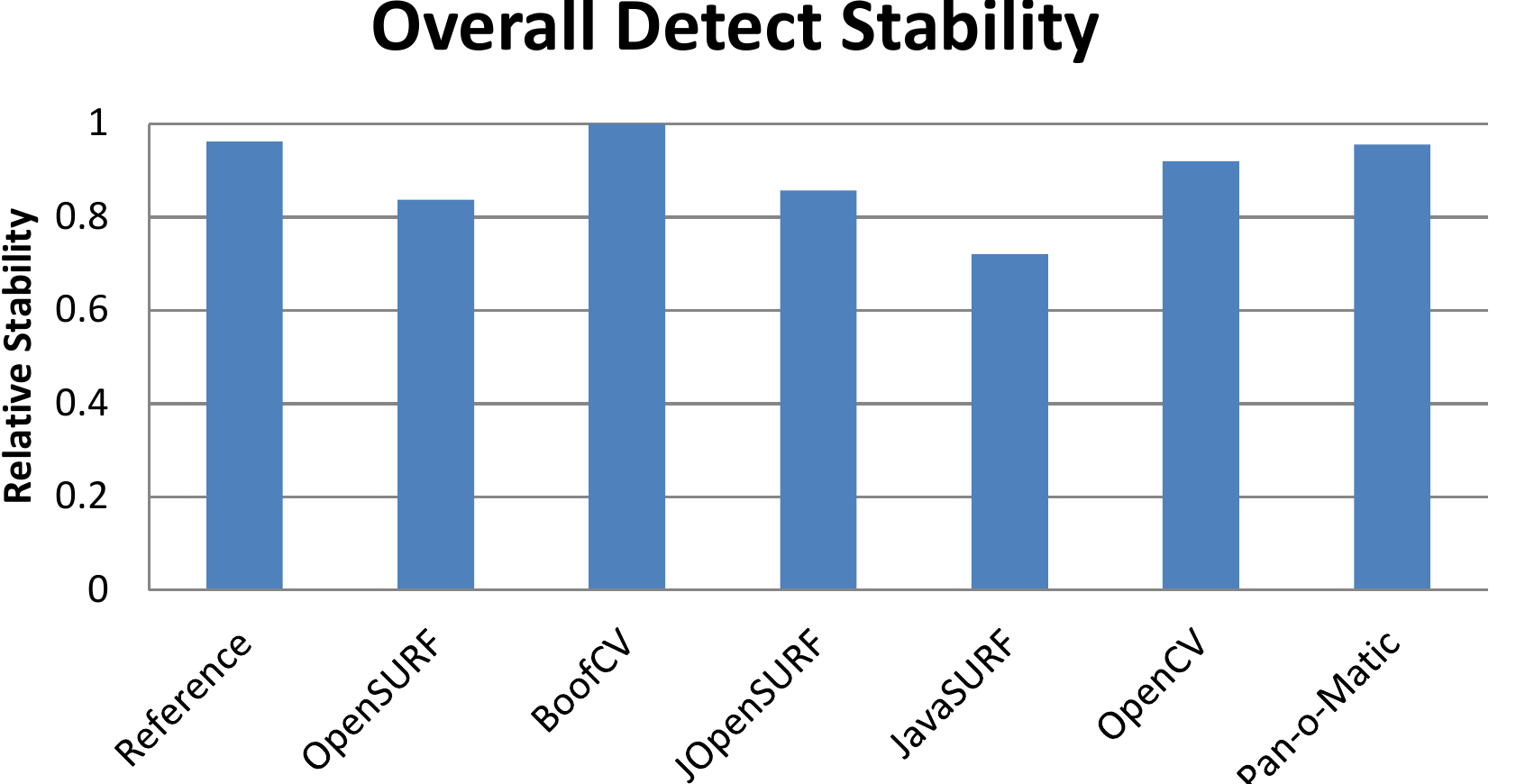}
\caption{\label{fig:overall_detector}Summary of detector stability using a modified repeatability measure across all image sequences.  Higher bars are better.  Tuning libraries to detect the same number of features across all images proved to be impossible; they are tuned to detect the same number of features in a single image and only unambiguous matches are considered.}
\end{figure}

Summary results for runtime performance, descriptor stability, and detector stability are shown in Figure \ref{fig:overall_runtime}, \ref{fig:overall_descriptor} , and \ref{fig:overall_detector} respectively.  Stability results for individual sequences have been omitted due to space constraints.  Descriptor performance has an approximate range of 40\% and detector performance has an approximate range of 25\%.  For runtime performance, the best implementation out performs the worst more than eight times.

BoofCV-M has the best descriptor stability by a small margin, followed by the reference library, and then Pan-o-Matic.  The same can be said for detector stability.  BoofCV-F is the fastest implementation, despite being written in Java.  The runners-up are OpenSURF and BoofCV-M, which have nearly the same runtime speed, but are two times slower than BoofCV-F.  A well-written C++ port of BoofCV-F is likely to run a minimum of two times faster.

Comparable overall results are found between \cite{Gossow2011} and this study, despite different procedures and metrics.  Both OpenCV and OpenSURF's implementations have been used to represent SURF's performance in recent literature \cite{Calonder10,Fischer2011,Luo2009}.  The version of those libraries used in this study did not exhibit behavior representative of the reference library for both describe and detect stability.

\section{Conclusions}

Important implementation details not covered or ambiguously described in the original SURF paper have been discussed.  To resolve these ambiguities, general techniques for enforcing the smoothness rule are defined and applied to SURF.  Best practices for maximizing stability and runtime speed were described in detail.  In addition, it was shown that performance can be improved by slightly modifying the original algorithm.

To highlight the importance of these issues, a performance study of eight SURF implementations was done.  Based on the results of this study, it is recommended that the reference library, Pan-o-Matic or BoofCV be used to represent SURF's descriptive abilities.  

Through minor modifications, it is possible to trade stability for speed, as was shown with BoofCV's two implementations.  It is still possible to generate large improvements in runtime speed without resorting to hardware specific implementations.

\bibliographystyle{plain}
\bibliography{surf_compare}
\end{document}